\documentclass[conference]{IEEEtran}

\usepackage{amsmath}
\usepackage[algo2e]{algorithm2e}

\usepackage{xcolor}
\usepackage[noend]{algpseudocode}
\usepackage{algorithm}
\usepackage{graphicx}
\usepackage{subcaption}

\begin{document}
\IEEEoverridecommandlockouts
\IEEEpubid{\makebox[\columnwidth]{979-8-3503-3959-8/23/\$31.00 \copyright 2023 IEEE \hfill} \hspace{\columnsep}\makebox[\columnwidth]{ }}


%
\title{Improved Block Merging for 3D Point Cloud Instance Segmentation}

\author{\IEEEauthorblockN{Leon Denis, Remco Royen and Adrian Munteanu}
\IEEEauthorblockA{Vrije Universiteit Brussel, Departement of Electronics and Informatics, 1050 Brussels, Belgium\\
Email: leon.denis@vub.be, remco.royen@vub.be, adrian.munteanu@vub.be}
}

\maketitle

\begin{abstract}
This paper proposes a novel block merging algorithm suitable for any block-based 3D instance segmentation technique. The proposed work improves over the state-of-the-art by allowing wrongly labelled points of already processed blocks to be corrected through label propagation. By doing so, instance overlap between blocks is not anymore necessary to produce the desirable results, which is the main limitation of the current art. Our experiments show that the proposed block merging algorithm significantly and consistently improves the obtained accuracy for all evaluation metrics employed in literature, regardless of the underlying network architecture.

\end{abstract}


%
\IEEEpeerreviewmaketitle

\section{Introduction}
Interest in 3D data is growing at a record pace. This is largely fuelled by the advancements in 3D data capturing technologies in combination with the AI breakthrough seen in recent years. One particularly important aspect related to this field is 3D scene understanding. Domains such as virtual reality, autonomous driving, and drone exploration require a deep understanding of the captured 3D scene. This includes semantic segmentation, which classifies the class of each object in the 3D point cloud, and instance segmentation, which performs semantic segmentation in combination with instance labelling for each point. 

The domain of 3D scene instance segmentation can be subdivided in two different groups, each targeting a different application. Belonging to the first group, one can distinguish the neural networks operating on the full scene at once. Those \textit{full-scene methods} are particularly useful for non-real-time applications where general scene understanding is not necessary during capturing or when enough resources are available during runtime. Inspection of office buildings can serve as an example. Important full-scene methods include \cite{jiang2020pointgroup, han2020occuseg, he2021dyco3d, chen2021hierarchical, vu2022softgroup}. In the first work, a network named PointGroup \cite{jiang2020pointgroup} learns offsets that aid a point clustering algorithm by increasing the intra-instance distances. Redundant clusters are subsequently removed by non-maximum suppression and a dedicated network that evaluates each obtained instance candidate. In DyCo3D \cite{he2021dyco3d} dynamical convolutions are employed to ultimately achieve instance segmentation. In a later work, a hierarchical aggregation is proposed to progressively generate instance proposals \cite{chen2021hierarchical}. SoftGroup \cite{vu2022softgroup} further improves upon this work and introduces soft semantic labelling prior to clustering.

The second group consist of \textit{block-based methods}. Unlike their full-scene counterparts, they operate on parts of the scene, and thus, are able to analyse 3D environments during point cloud capturing. This is particularly useful for applications such as autonomous driving and drone exploration, which typically need to interpret the currently captured scene in order to navigate. Recent works in this field include ~\cite{wang2018sgpn, wang2019associatively, zhao2020jsnet, zhang2021point, chen2022jspnet, elich20193d, pham2019jsis3d, he2020instance}. Similar as the full-scene methods, they often rely on clustering techniques. SGPN \cite{wang2018sgpn}, for example, computes a similarity matrix which is subsequently used for grouping similar points to one particular instance. ASIS \cite{wang2019associatively}, JSNet \cite{zhao2020jsnet} and JSPNet \cite{chen2022jspnet} jointly perform semantic and instance segmentation through specifically designed modules connecting both branches of the network. The resulting feature vector once again serves as input for a clustering algorithm. In \cite{zhang2021point}, each point is represented as a tri-variate normal distribution and a novel loss function is employed for clustering. Lastly, MP-Net \cite{he2020learning} proposes a memory-augmented network which learns and memorizes feature prototypes for scene analysis. 

As mentioned previously, all aforementioned methods belonging to this group process the scene in a block-based manner. Full scene analysis is obtained by combining the information of each block using a merging algorithm. As will be clarified in the remainder of this paper, the employed algorithm can greatly affect the obtained results. Yet, to the best of our knowledge, only a single block merging algorithm to date exists, which we will henceforth refer to as 'BlockMerging v1' \cite{wang2018sgpn}. It is used by \emph{all} aforementioned block-based segmentation techniques. The proposed work has one major contribution. It proposes a novel block merging algorithm that vastly improves over the current art. Unlike the original method, the improved algorithm considers historical information and allows already labelled points to be altered to correct wrongly labelled instances from past blocks. Consequently, the improved block merging algorithm consistently and simultaneously improves performance with respect to multiple evaluation metrics, regardless of the employed architecture.


 

\section{BlockMerging v1}
Block-based approaches for 3D scene instance segmentation partition a given scene in blocks prior to processing. These blocks serve as input for the block merging algorithm. The area of the block size is commonly chosen as $1m \times 1m$, which corresponds to the original proposal of \cite{wang2018sgpn} and which is also retained in this work. We note that blocks are constructed based on the ground level. Infinite height is assumed. We define the $i^{th}$ block as $B_i$ from the set of blocks $B$. In addition to block-based partitioning, the input scene is also voxelized. The size of each cell is chosen such that the input scene fits a $400 \times 400 \times 400$ grid. The voxelization, thus, operates on all 3 spatial dimensions, unlike block partitioning. Each point of the scene is assigned to the voxel, which spatially embeds said point. We mathematically express this as:
\begin{equation}
c_{j} = C(p_n),
\end{equation}
with $c_{j}$ the corresponding voxel (or cell) of input point $p_n$ and $C$ the voxelization operation. The general idea behind block merging is defining a label for each cell based on the predicted labels of its population. The method starts by defining for each cell $c_j$ of the first block $B_1$ the label $k$ selected from the set of instances identifiers $K$ where 
\begin{equation}
\sum_{n=1}^{N}[\sigma(p_{j,n})=k]
\label{eq:first_block}
\end{equation}
is maximal. $\sigma$ is the inference operator of the neural network operating on the $n^{th}$ input point $p_{j,n}$ belonging to cell $c_j$. We denote $l_{c_j} = k$ as the label belonging to $c_j$. Next, blocks are processed in an overlapping manner using a sliding window with snake pattern and stride 0.5 meter, as shown in Figure \ref{fig:limitation}. We note that other scanning methods can improve performance. For each block, all instances $I_m$ are traversed and the labels $l_{c_j}$ are assigned to the cells $C_{j,m}$ belonging to $I_m$ if the number of occurrences of $l_{c_j}$ is larger than a given threshold $\tau_1$. If not, the cells $C_{j,m}$ are labelled with a new monotonically increasing grouping identifier, which represents a new instance. This processes is repeated until all blocks are processed after which all $p_{j,n}$ in $c_j$ are labelled as $l_{c_j}$.

\subsection{Limitations}
\begin{figure}[b]
  \centering
   \includegraphics[width=0.66\linewidth]{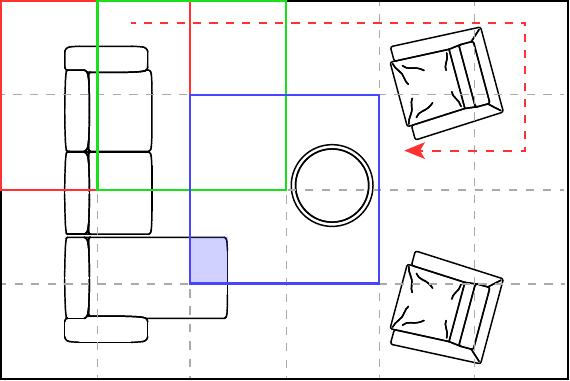}
   \caption{Top down view of a room divided in blocks. The portion of the sofa marked in blue will be wrongly labelled using current block merging techniques \cite{wang2018sgpn} as no overlap is present.}
   \label{fig:limitation}
\end{figure}

Though performing well in most cases, the original block merging algorithm has one major limitation \cite{wang2018sgpn}. The underlying snake pattern does not ensure instance overlap when processing blocks. That is, it is possible that only not-yet-labelled points of a particular instance identified in the past are present in a given block. This, combined with the monotonically increasing grouping identifier results in the failure of grouping numerous instances. This is demonstrated by Figure \ref{fig:limitation}. The dashed lines indicate the position of all blocks and are drawn with 0.5 meter offset. Note that the blocks themselves are 1m\textsuperscript{2}. The first and second blocks are depicted in red and green, respectively. They process the top part of the sofa, and label the associated points with a given instance number. The problem arises when processing the 8\textsuperscript{th} block, depicted in blue. That particular block is only able to process the portion of the sofa that is marked. As there is no overlap with other points associated with the sofa, the currently processed points will be wrongly labelled as a new instance.

\section{Proposed method}
To alleviate this issue, we propose a novel block merging algorithm that is able to alter past labelled cells while processing any given block. This is done by allowing the merging of two or more different labels when object confusion is detected as more information of different blocks becomes available. The pseudocode of our algorithm is shown in Algorithm \ref{alg:hs}. Red lines indicate the lines of code introducing the new concepts. We will refer to this method as 'BlockMerging v2'.  Instead of permanently labelling  processed cells, labels $l_{c_{j,m}}$ associated with cells from instance $I_m$ are stored in the list $L_{I_m}$. Note that the overlapping windows allow multiple $l_{c_{j,m}}$ to be associated with the same instance $I_m$, which is exactly the root of the problem, resulting in subdivided instances. Once a block has been processed, labels are propagated on a per-instance basis using the instances resulting from the block-based prediction. That is, should there exist multiple $l_{c_{j,m}}$ for any given instance $I_m$, then the labels in $L_{I_m}$ are redefined as $l_{I_{max}}$ such that
\begin{equation}
\sum_{m=1}^{M}[l_{c_{j,m}}=L_{I_{max}}]
\label{eq:proposed}
\end{equation}
is maximal. In other words, the ambiguity of the labels associated with a given instance is removed by labelling each cell $C_{j,m}$ with the most occuring label $L_{I_{max}}$. We note that the labels $l_{c_{j,m}}$ are altered on a global level while instances $I_m$ are processed in a sequential, but not necessarily successive, order. This allows propagating labels farther than a single block-size, ultimately leading to full-scene label propagation.  



\begin{algorithm}
    \caption{- BlockMerging v2}
    \label{alg:hs}
    \DontPrintSemicolon
    \KwIn{Blocks: $B$, Instances per block: $I$, Voxel grid: $V$}
    \KwOut{Point instance labels for the whole scene: L}
    {Initialize all cells $c_j$ of $V$ as $-1$\\}
    {group $\gets 0$ \\ }    
    \For{every block $B_i$}
    {
        \eIf {$B_i$ is the first block}
        {
           \For{every point $p_{n}$ in $B_i$ belonging to $I_m$}
           {
            Define $c_j = C(p_n)$ \\
            $l_{c_j} \gets \sigma(p_{n})$ \\
            Add $l_{c_j}$ to the instance mapping table $H_{I_m}$\\
            \If{ $group <= l_{c_j}$}
            {
                $group \gets$ \text{$l_{c_j} + 1$}
            }
           }
        }
        {
            \For{every instance $I_{m}$}
            {
                {Define cells ${c_j}$ \text{which embed points of} $I_m$ \\}
                {Define $l'_{I_{max}}$ as the mode of $I_m$ with cells not $-1$}
                \If{ $l_{c_j}$ is not -1}
                {                    
                    \eIf{frequency of $l'_{I_{max}} < \tau_1$}
                    {
                        {$l_{c_j} \gets$ \text{$group$} \\
                        $group \gets$ \text{$group + 1$}}
                    }
                    {
                    ${l_{c_j}} \gets l'_{I_{max}}$
                    }
                Add $l_{c_j}$ to the instance mapping table $H_{I_m}$  
                }
            }
            \For{every instance $I_{m}$}
            {                
                \For{every label $l_{I_j}$ belonging to $I_m$}
                {
                    Use $H_I$ to define $l_{I_{max}}$ as the mode of $I_m$ \\
                    \If{the frequency of $l_{I_j} > \tau_2$}
                    {
                       $l_{I_j}$ = $l_{I_{max}}$
                    }
                }
            }
        }
        \For{every point $p_n$ in the whole scene}
        {
            {Define $c_j$ as $C(p_n)$}\\
            {$L_{p_n} \gets l_{c_j}$}
        }
    }
\end{algorithm}

To provide more control of the merging process, we also introduce a control parameter $\tau_2$, which prevents merging when the number of cells with a given label is above a threshold for a particular instance. Larger thresholds allow more label ambiguity and thus, instance partitioning, indirectly resulting in smaller instances being detected, hence, improving recall. On the other hand, a smaller $\tau_2$ will faster enforce label merging, which in turn increases overall precision. Using the control parameter $\tau_2$, one is, thus, able to  fine-tune the overall inference results and balance recall and precision to a limited, but nonetheless useful, extend. More importantly, empirical results, and an ablation study presented later in this paper, show that recall and precision can both be improved simultaneously when using the proposed method compared to the original block merging algorithm \cite{wang2018sgpn}.

\section{Experimental assessments}
\begin{figure}
  \centering
   \includegraphics[width=0.88\linewidth]{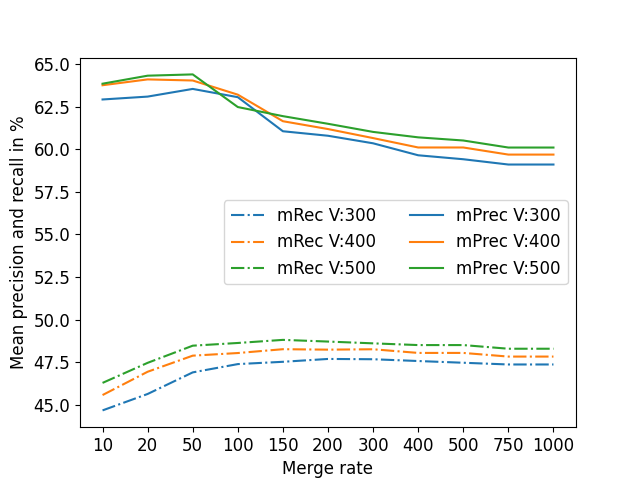}
   \caption{The effect of the parameter $\tau_2$ with respect to the mean precision and mean recall for different voxel sizes.}
   \label{fig:merge_rates}
\end{figure}

As a first experiment, we demonstrate the influence of the threshold $\tau_2$ with respect to the obtained performance, which is shown in Figure \ref{fig:merge_rates}. For this specific experiment, the publicly available model of JSNet was employed \cite{zhao2020jsnet}. From the figure, one can notice similar behaviour for all voxel sizes. Importantly, mean precision and recall \emph{both} benefit at low thresholds. As expected, degraded performance is noticeable when choosing $\tau_2$ too large. It is also noteworthy that better overall accuracy is obtained with increasing resolution of the voxel grid. Peak performance is obtained with thresholds around 50 and 100. Similar observations were made for the mean coverage and mean weighted coverage.
In terms of network accuracy, we have compared the proposed method to the original BlockMerging algorithm of \cite{wang2018sgpn} using numerous instance segmentation architectures \cite{wang2019associatively,zhao2020jsnet, chen2022jspnet, jiang2020pointgroup, he2021dyco3d, chen2021hierarchical}. Besides block-based methods, we have also incorporated networks normally operating on full scenes. Their models have been retrained specifically for blocks. For the other networks, publicly available models were employed. The only exception is the most recent state-of-the-art block-based method of \cite{chen2022jspnet}, for which the results of the original work are reported. We were unable to apply the proposed block merging on the output of \cite{chen2022jspnet} as its source code is not made public fully. However, we report their results to allow making a comparison with the state-of-the-art in block-based instance segmentation.

Table \ref{tab:results} and Table \ref{tab:results2} summarize the findings of our experiments. For grid size $400^3$ a $\tau_2$ of 110 was employed. For $500^3$ $50$ was selected for all architectures except for ASIS \cite{wang2019associatively}, for which 110 was used. From the tables, it is clear that the proposed method significantly improves upon BlockMerging v1 \cite{wang2018sgpn}. This is consistently true for \emph{all} evaluated architectures and for \emph{all} metrics. It is also noteworthy that the obtained results for JSNet are slightly lower than reported in their original work \cite{zhao2020jsnet}. However, even so, employing BlockMerging v2 with JSNet significantly outperforms the state-of-the-art method of \cite{chen2022jspnet}, improving mean recall with 0.2\% and mean precision with a significant 3.5\% when taking the highest available values for both metrics into account. When using a grid of $500^3$ even better results are obtained, improving mean recall and precision with 0.5\% and 4.8\%, respectively. The best mean and weighted coverage are still produced by JSPNet. However, given the outcome of the current experiments, it is fair to assume better results would be obtained when utilising the proposed block merging for that architecture as well. 

Looking at individual improvements, compared to BlockMerging v1 \cite{wang2018sgpn}, the proposed method increases mean recall up to 2.8\%, and mean precision up to 7.2\%. For mean coverage and weighted coverage, the additional performance increase adds up to 1.8\% and 2.5\%, respectively. These results clearly indicate the benefits of the proposed block merging algorithm. 

\begin{figure*}
 \begin{minipage}{\textwidth}
 \begin{center}    
    \subfloat{ \includegraphics[width=0.22\textwidth]{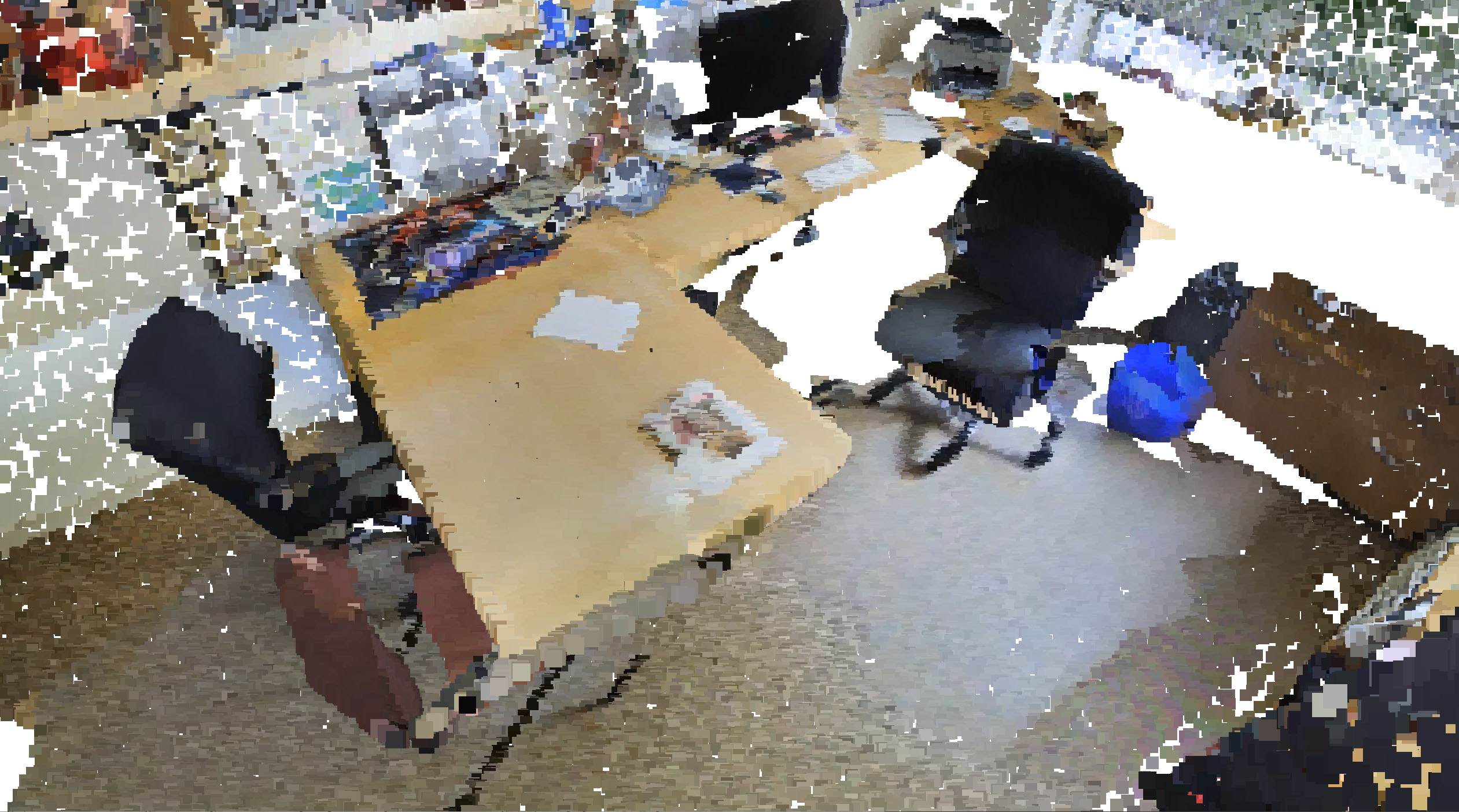}}
    \subfloat{ \includegraphics[width=0.22\textwidth]{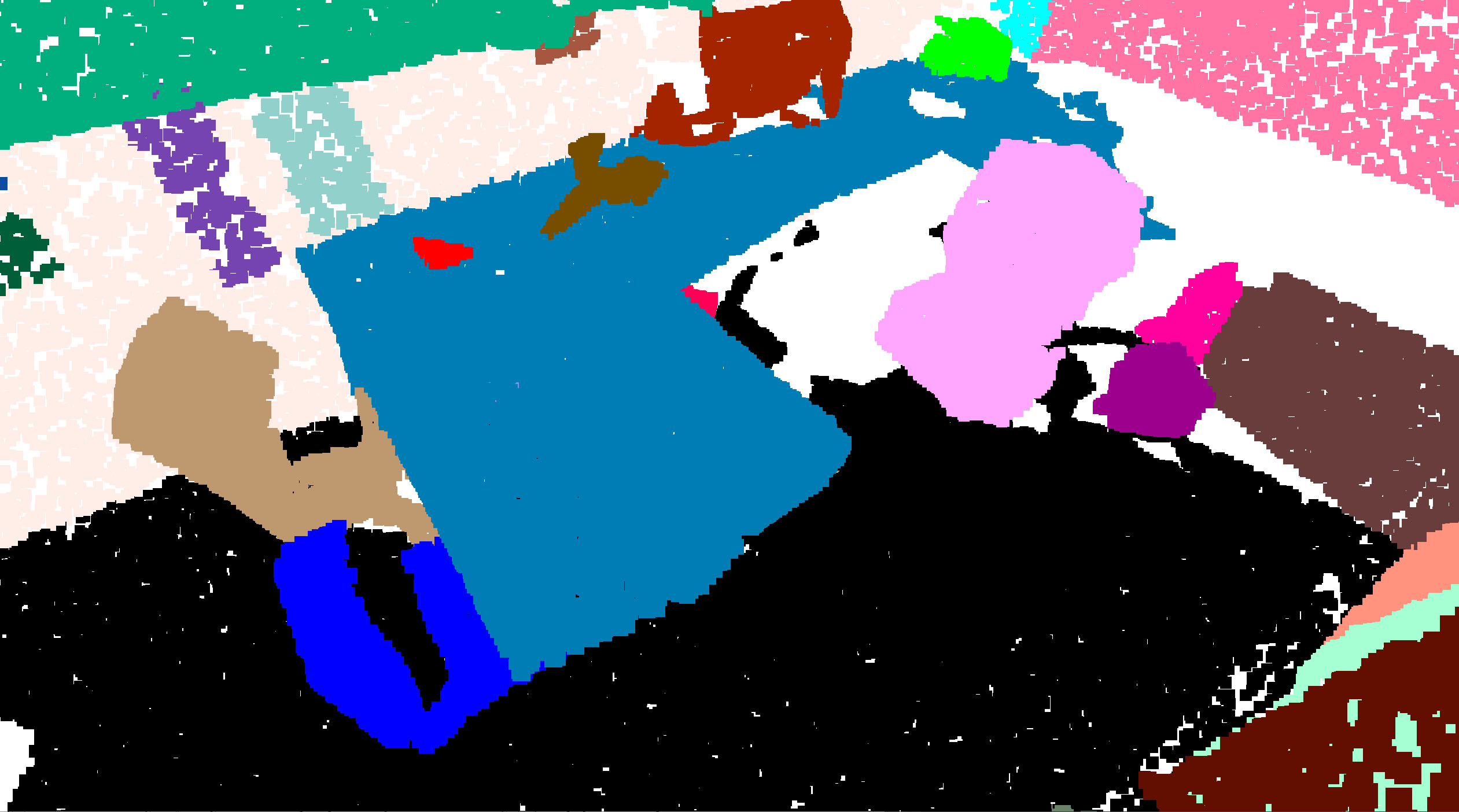}}
    \subfloat{ \includegraphics[width=0.22\textwidth]{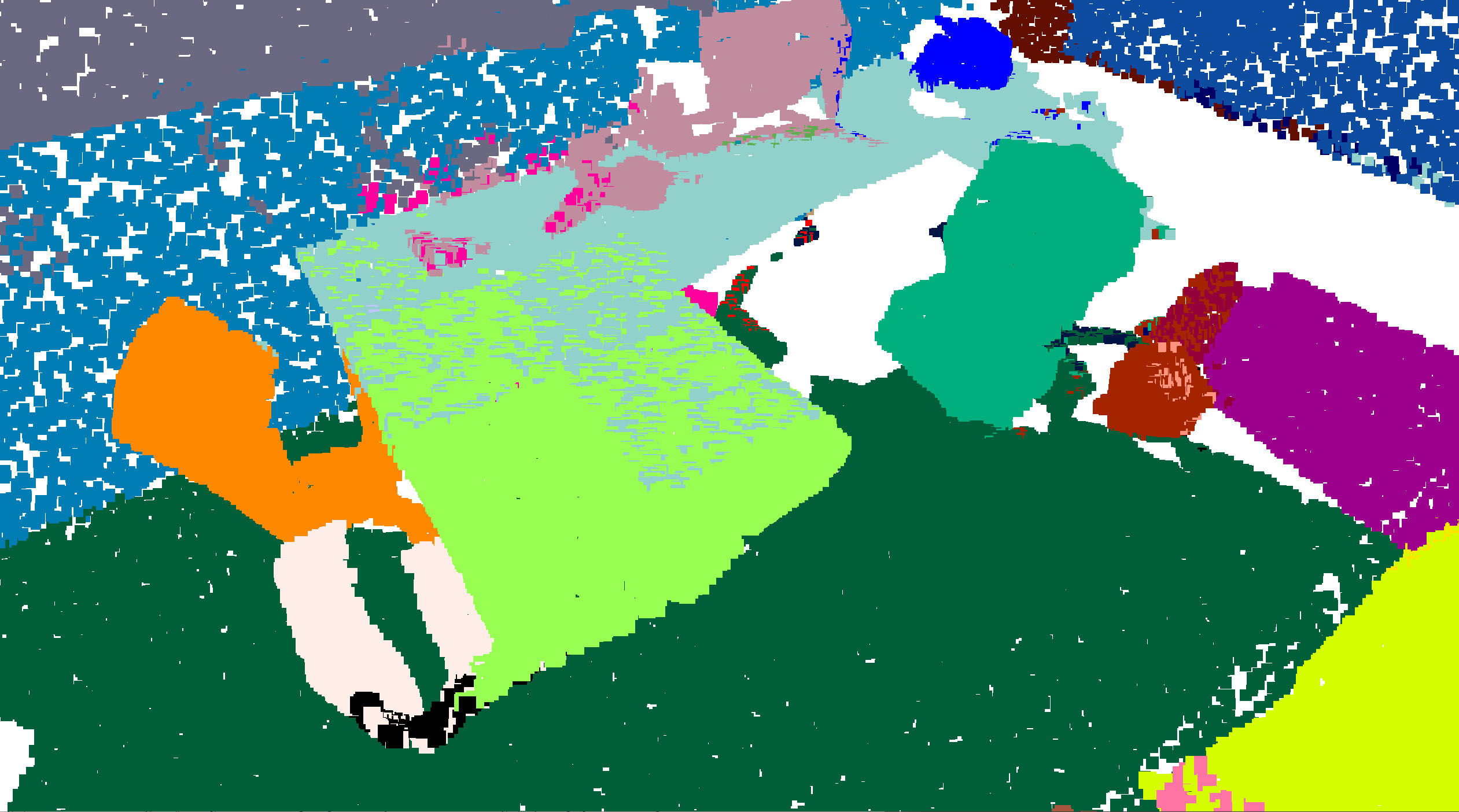}}
    \subfloat{ \includegraphics[width=0.22\textwidth]{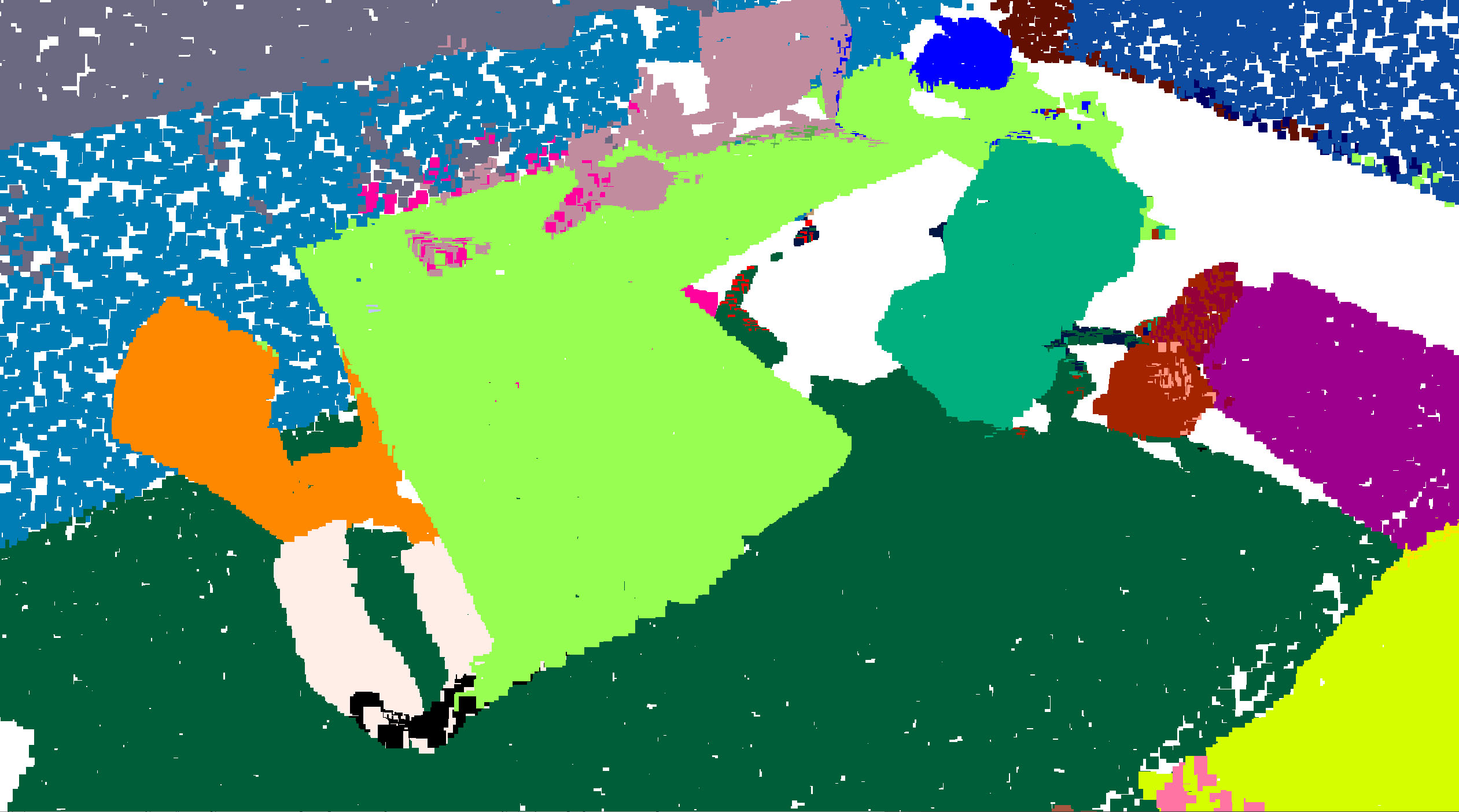}}
    \par\smallskip
  \end{center}
 \end{minipage}
 \begin{minipage}{\textwidth}
 \begin{center}    
  \subfloat{ \includegraphics[width=0.22\textwidth]{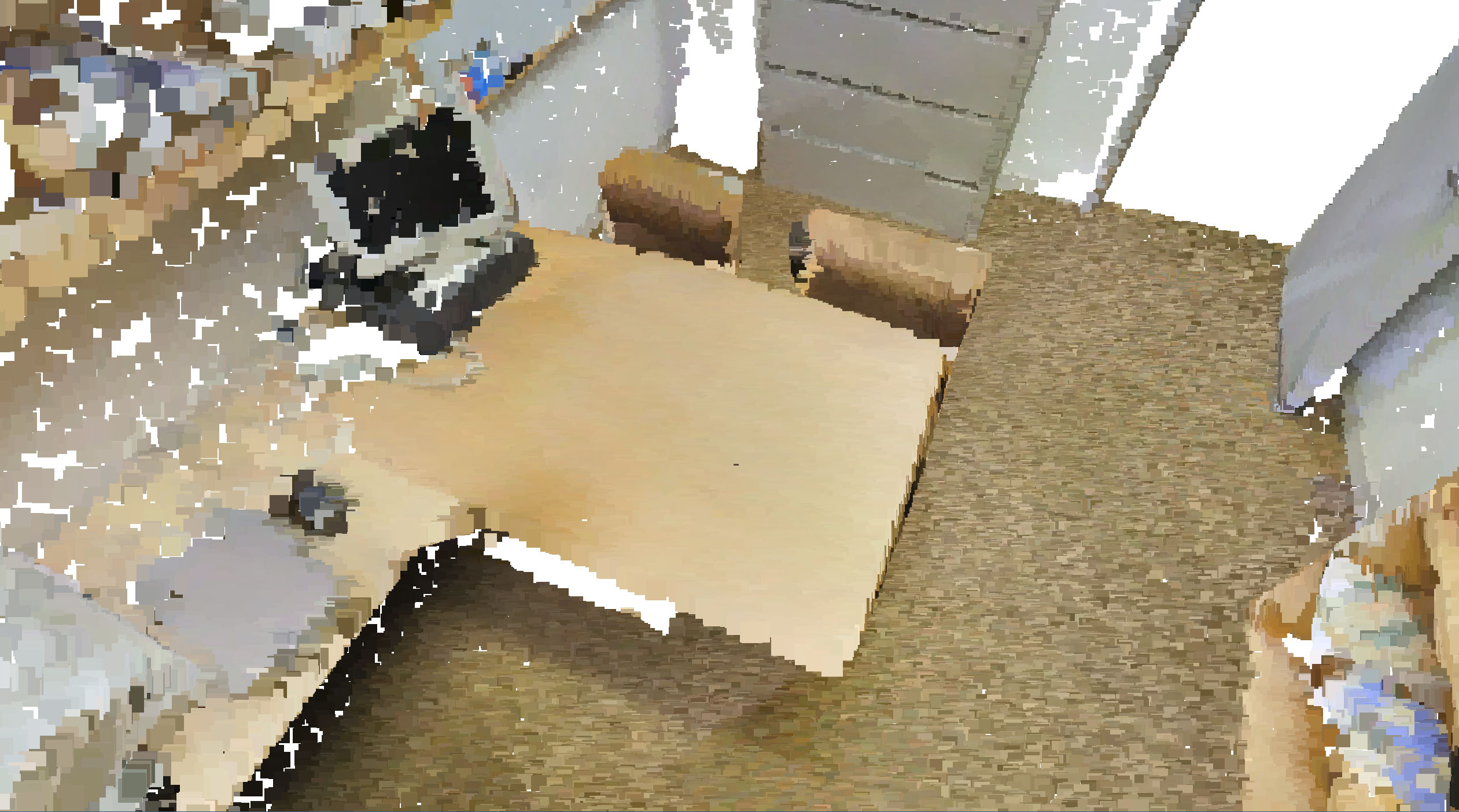}}
    \subfloat{ \includegraphics[width=0.22\textwidth]{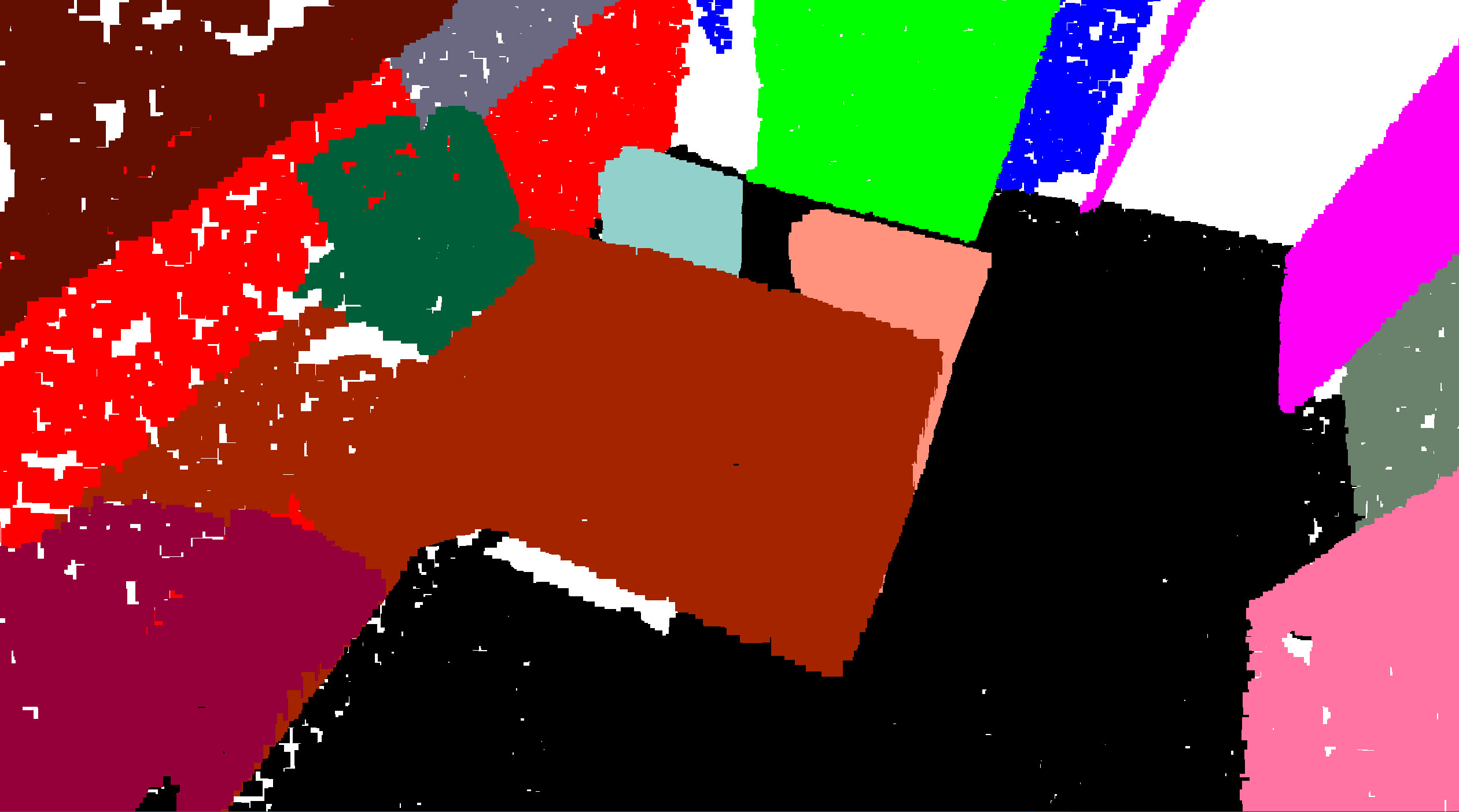}}
    \subfloat{ \includegraphics[width=0.22\textwidth]{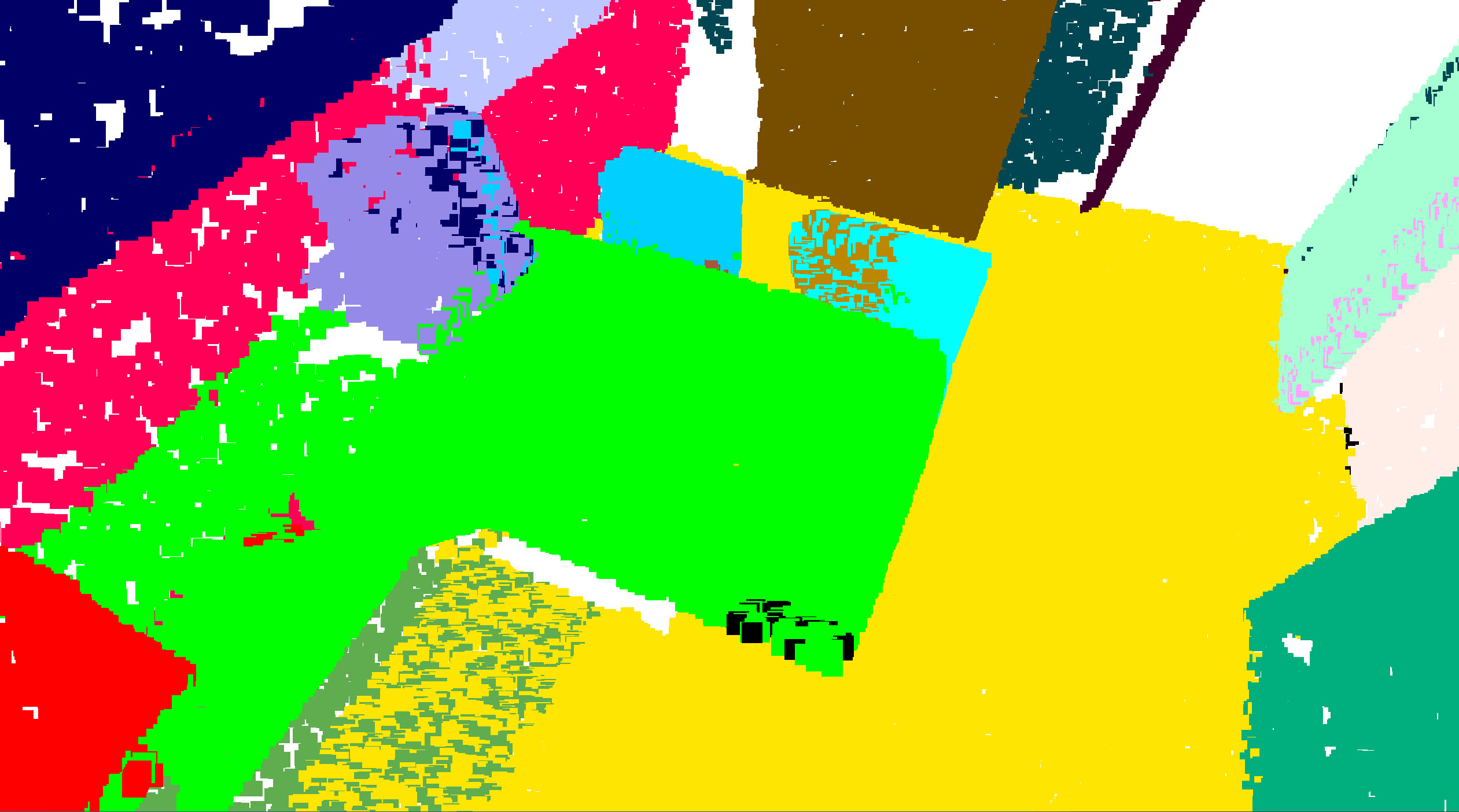}}
    \subfloat{\includegraphics[width=0.22\textwidth]{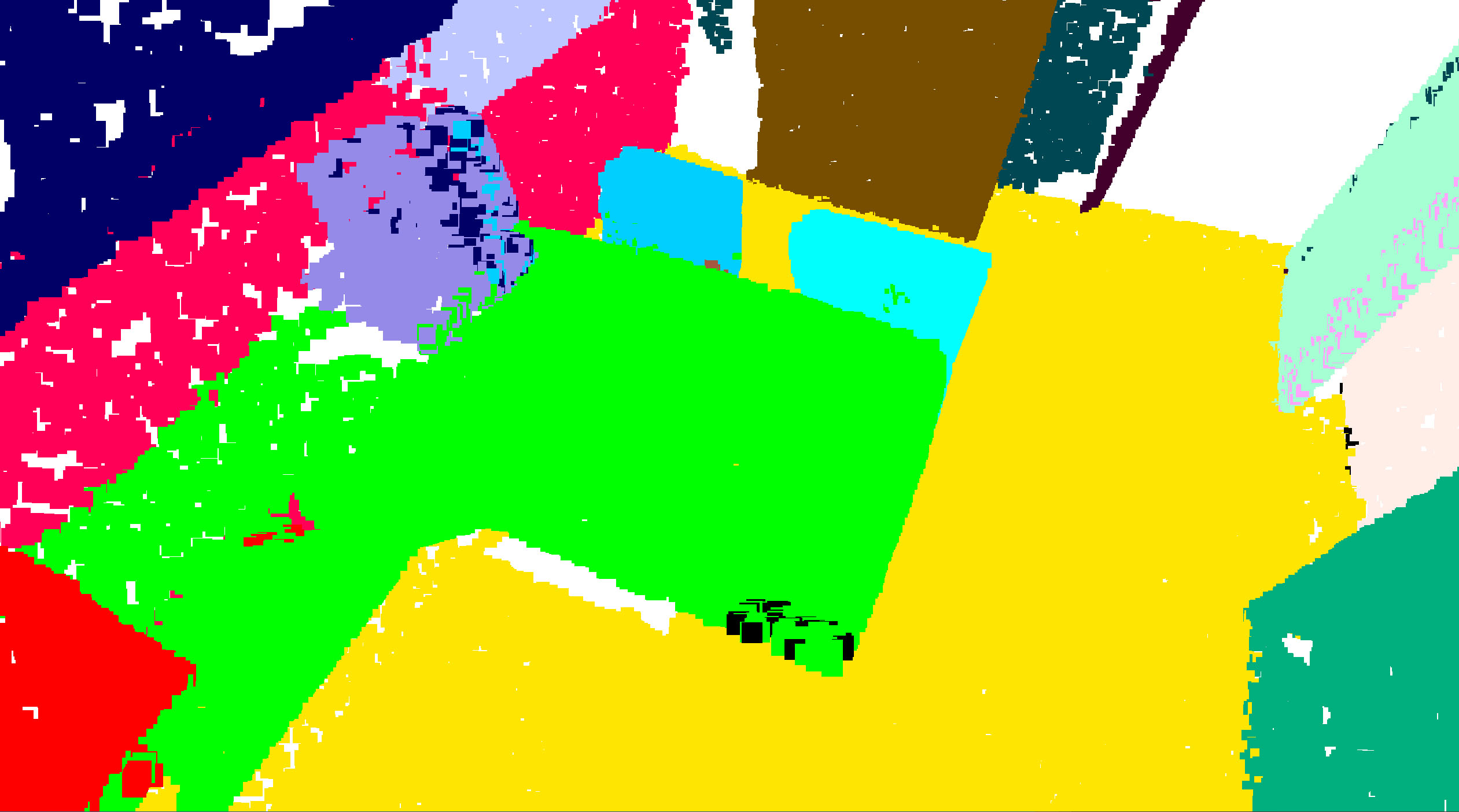}}    
  \caption{Results for grid resolution $400^3$. From left to right: input, ground truth, BlockMerging v1  \cite{wang2018sgpn}, proposed.}
  \label{fig:visual_400}
  \end{center}
 \end{minipage}
 \begin{minipage}{\textwidth}
 \begin{center}    
    \subfloat{ \includegraphics[width=0.22\textwidth]{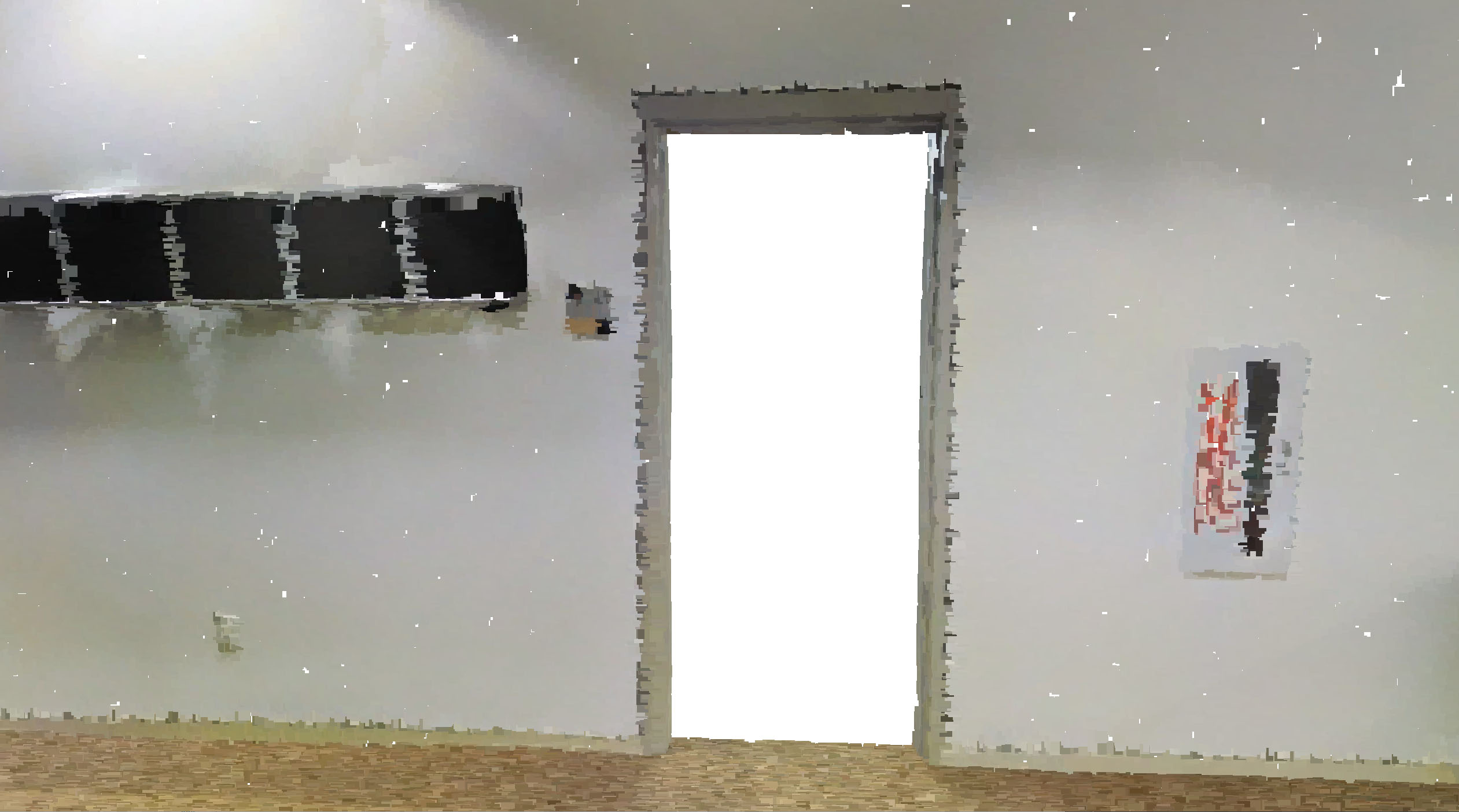}}
    \subfloat{ \includegraphics[width=0.22\textwidth]{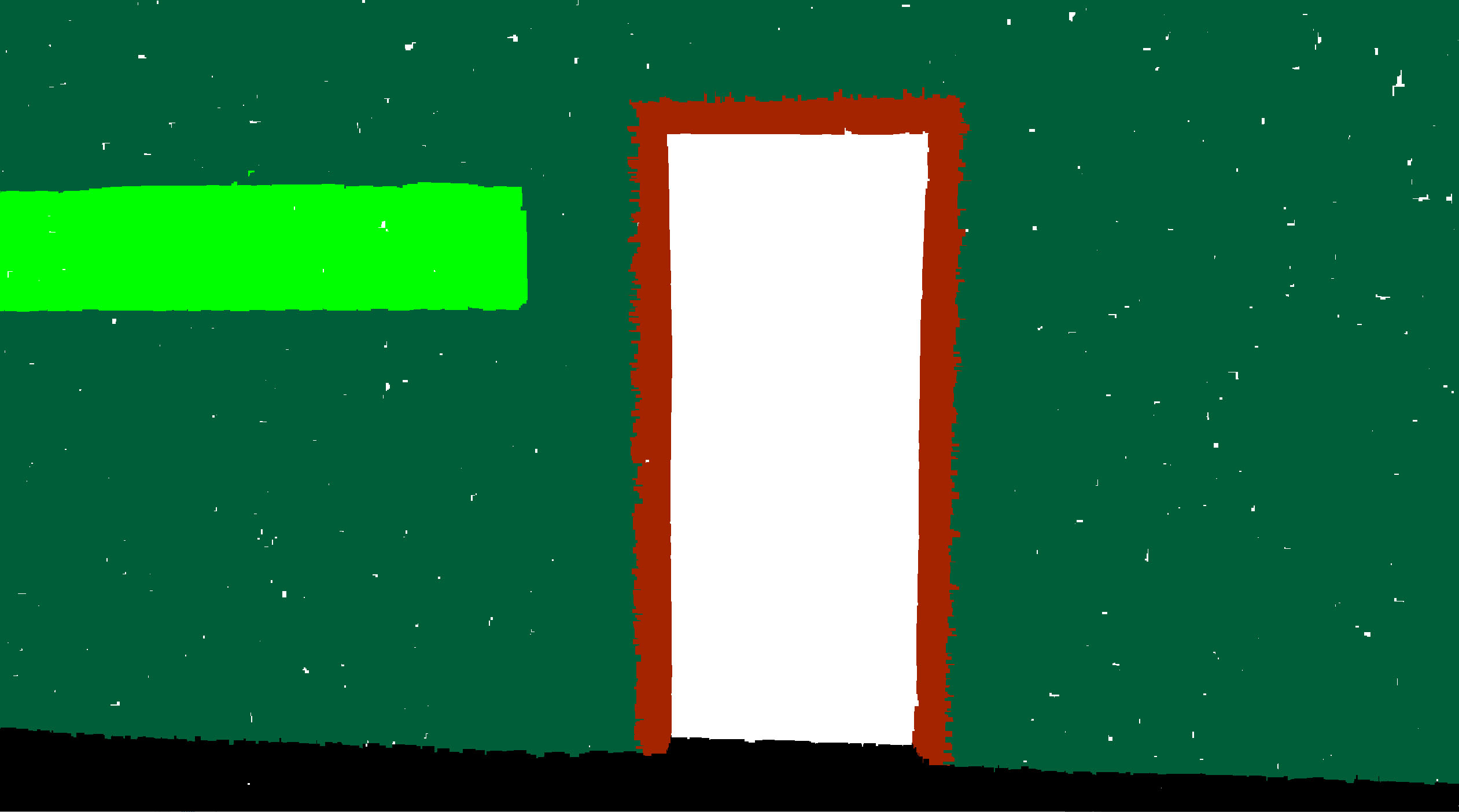}}
    \subfloat{ \includegraphics[width=0.22\textwidth]{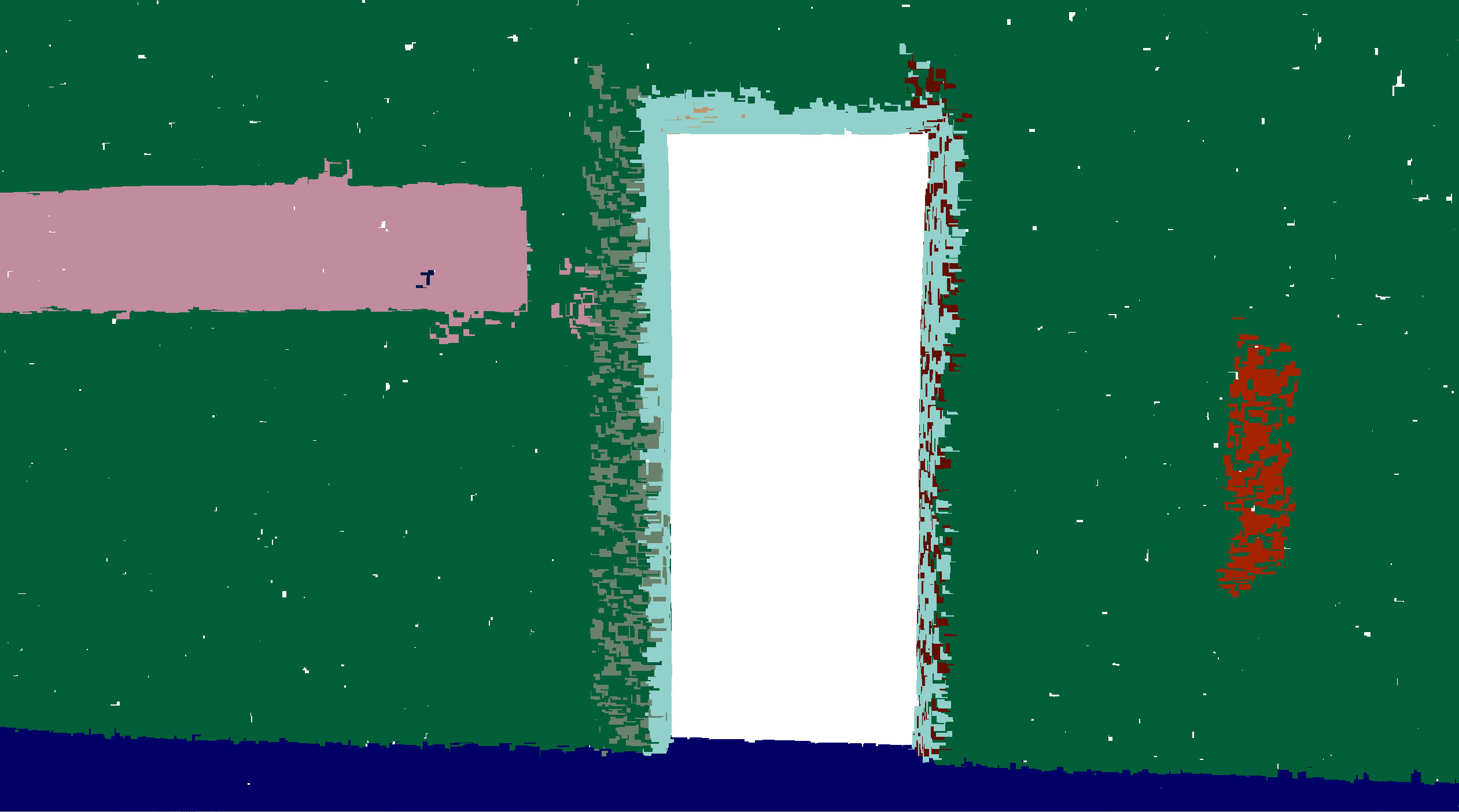}}
    \subfloat{ \includegraphics[width=0.22\textwidth]{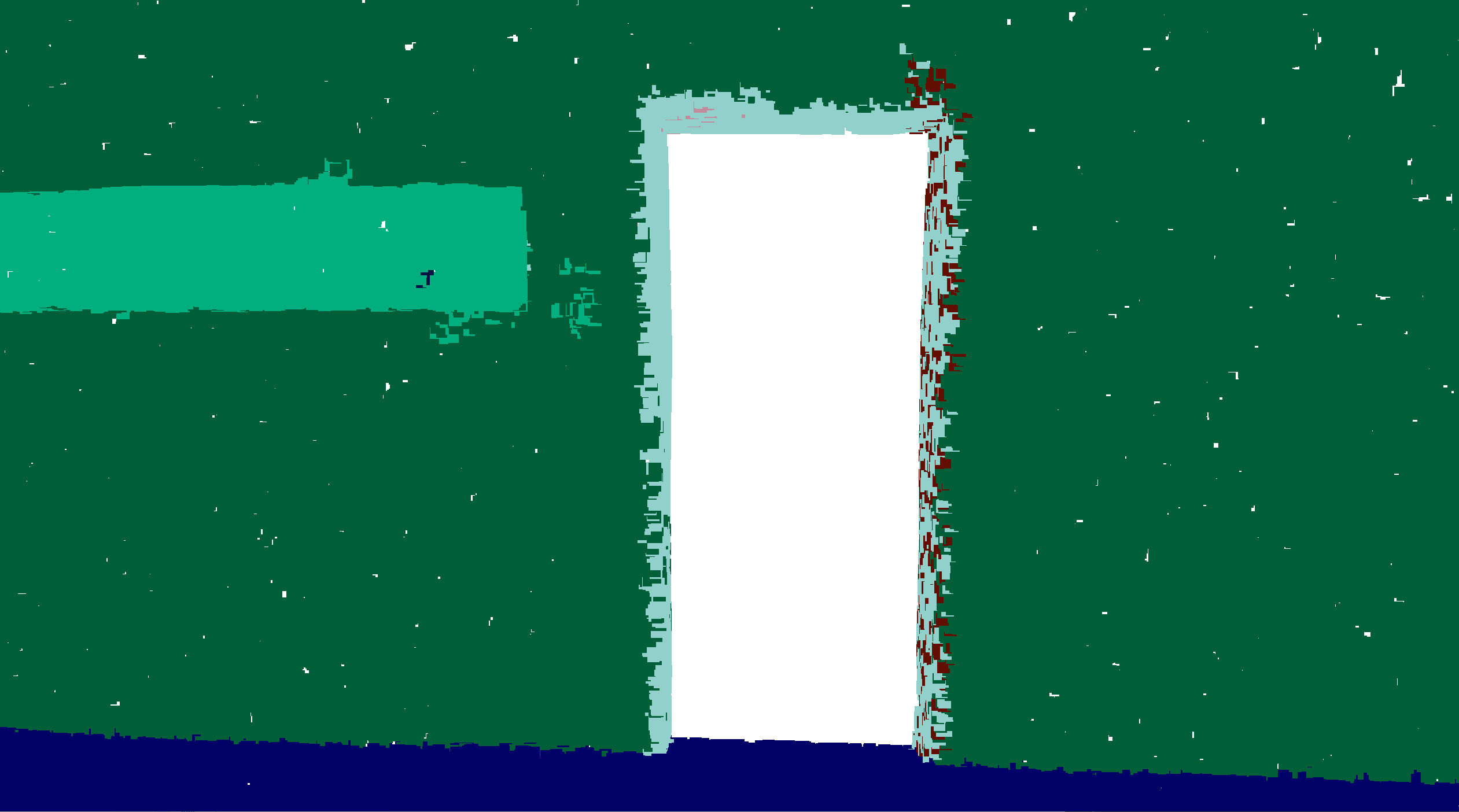}}
  \par\smallskip
  \end{center}
 \end{minipage}
 \begin{minipage}{\textwidth}
 \begin{center}    
  \subfloat{ \includegraphics[width=0.22\textwidth]{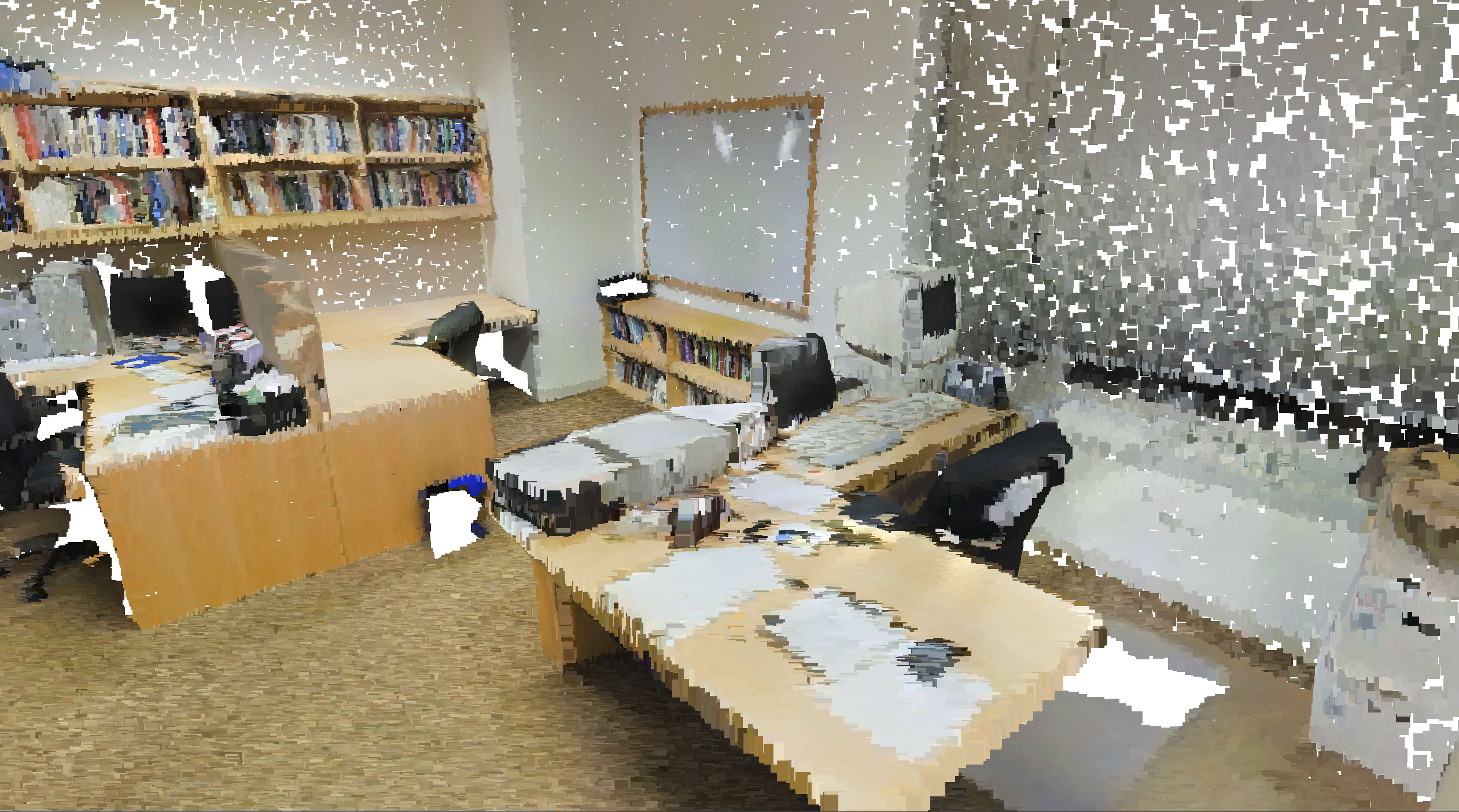}}
    \subfloat{ \includegraphics[width=0.22\textwidth]{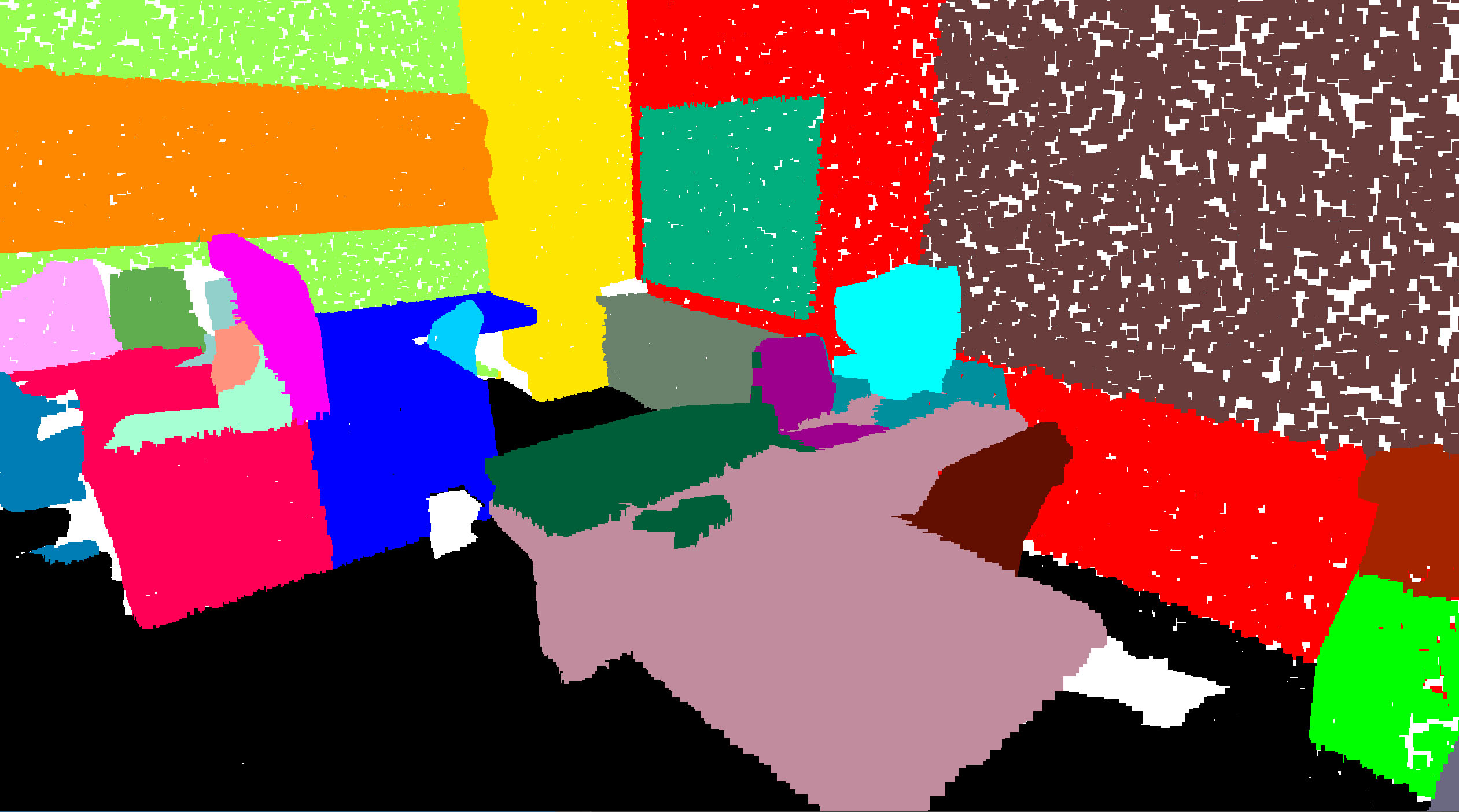}}
    \subfloat{ \includegraphics[width=0.22\textwidth]{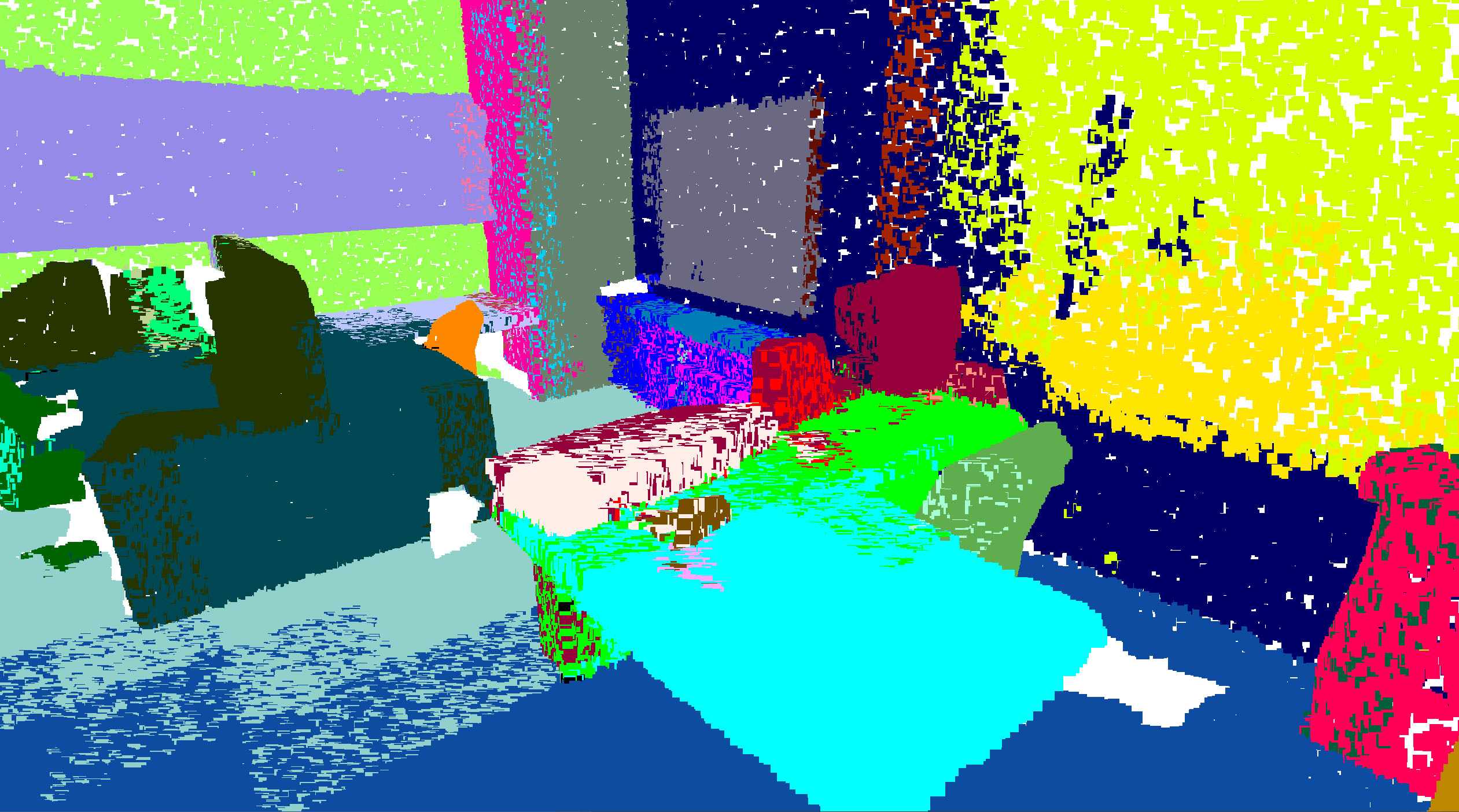}}
    \subfloat{ \includegraphics[width=0.22\textwidth]{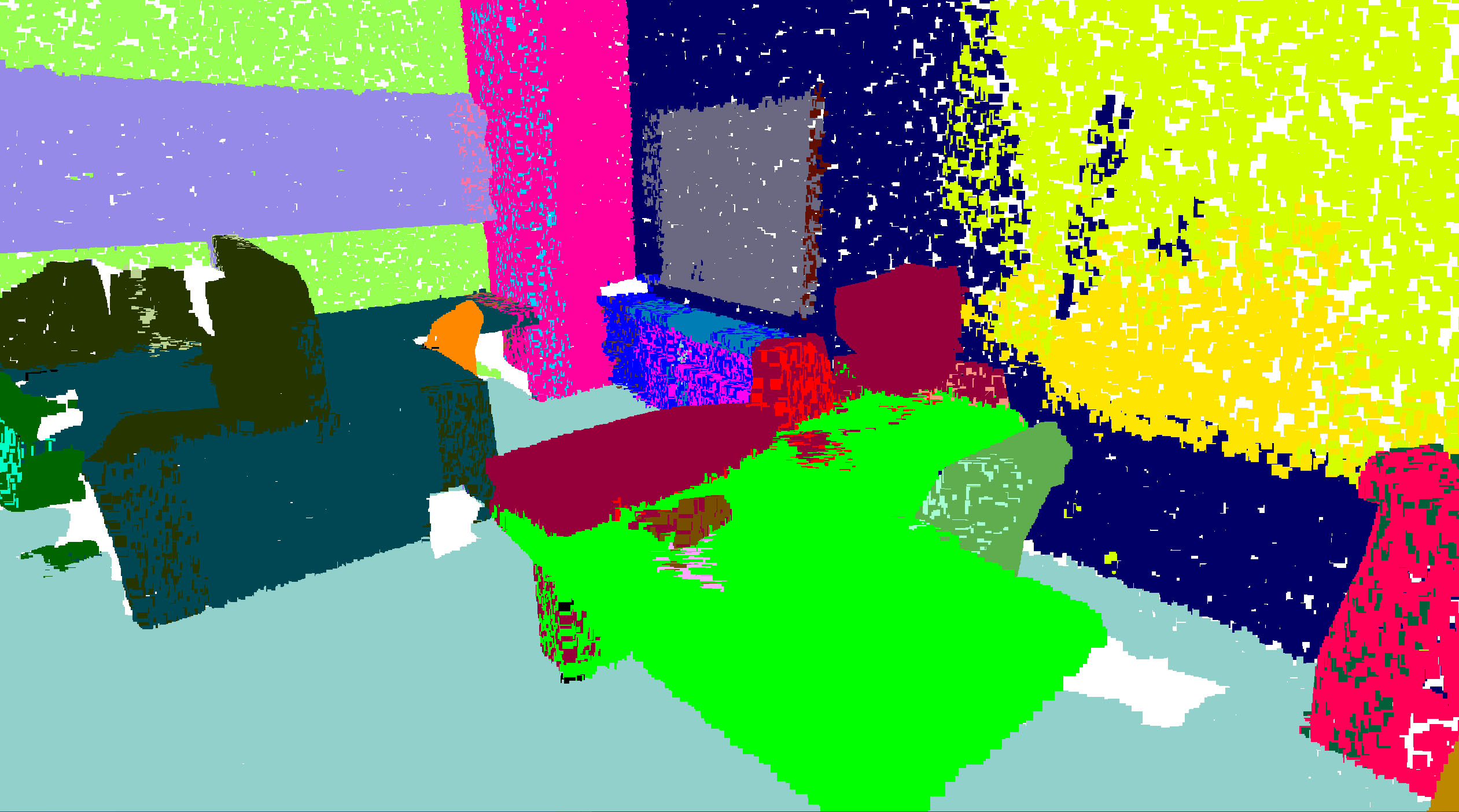}}
  \caption{Visual results for grid size $500^3$.}
  \label{fig:visual_500}
  \end{center}
 \end{minipage}
\end{figure*}

Visual results for grid sizes 400\textsuperscript{3} and 500\textsuperscript{3} are shown in Figure \ref{fig:visual_400} and Figure \ref{fig:visual_500}, respectively. The figures demonstrate the improved accuracy when using the proposed method. Less object confusion is noticed, and instances that were wrongly subdivided in multiple instances are now correctly fused. This is especially noticeable for the bottom row in Figure \ref{fig:visual_500}. Though some object confusion is still present, the proposed method clearly and greatly improves the overall segmentation results.  

Regarding the time complexity, the newly introduced functionality increased the execution time with roughly 29\%, on a laptop with an Intel i9-12900h.


\begin{table*}[t]
\begin{center}
\caption{Instance segmentation results on S3DIS-blocks for Area-5 using the original  \cite{wang2018sgpn} and proposed block merging methods. The models denoted with * are full-scene methods retrained on blocks.}
\begin{tabular}{|c|c cc c|c cc c|c cc c|} 
    \hline
    \multicolumn{1}{|c|}{} & \multicolumn{4}{c|}{BlockMerging v1  \cite{wang2018sgpn}} & \multicolumn{4}{c|}{Proposed} & \multicolumn{4}{c|}{Delta}\\
    \cline{2-13}
    Method & mRec & mPrec &mCov &mWCov & mRec & mPrec &mCov &mWCov& mRec & mPrec &mCov &mWCov\\ 
    \hline
    ASIS~\cite{wang2019associatively}              & 35.2 & 46.4 & 34.7 & 41.2 & 35.3 & 49.0 & 35.0 & 42.2 & 0.1 & 2.6 & 0.3 & 1.0 \\ 
    JSNet~\cite{zhao2020jsnet}              & 47.8 & 59.7 & 44.0 & 49.6 & \bf{48.2} &  \bf{63.2} & 44.6 & 50.6 & 0.4 & 3.5 & 0.6 & 1.0\\ 
    JSPNet~\cite{chen2022jspnet}              & 48.0 & 59.6 & \bf{50.7} & \bf{53.5} & - & - & - & - & - & - & - & -\\ 
    PointGroup*~\cite{jiang2020pointgroup}  & 45.4 & 48.1 & 41.6 & 46.9 & 46.9 & 53.4 & 42.6 & 48.6 & 1.5 & 5.3 & 1.0 & 1.7\\ 
    DyCo3D*~\cite{he2021dyco3d}             & 47.2 & 55.2 & 43.8 & 49.1 & 47.6 & 57.9 & 44.6 & 50.5 & 0.4 & 2.7 & 0.8 & 1.4\\ 
    HAIS*~\cite{chen2021hierarchical}       & 42.9 & 48.3 & 39.6 & 45.3 & 44.7 & 52.9 & 41.0 & 47.5 & 1.8 & 4.6 & 1.4 & 2.2 \\     
    \hline
\end{tabular}

\label{tab:results}
\end{center}
\end{table*}

\begin{table*}[t]
\begin{center}
\caption{Same table as Table \ref{tab:results}, but for grid size $500 \times 500 \times 500$.}
\begin{tabular}{|c|c cc c|c cc c|c cc c|} 
    \hline
    \multicolumn{1}{|c|}{} & \multicolumn{4}{c|}{BlockMerging v1  \cite{wang2018sgpn}} & \multicolumn{4}{c|}{Proposed} & \multicolumn{4}{c|}{Delta}\\
    \cline{2-13}
    Method & mRec & mPrec &mCov &mWCov & mRec & mPrec &mCov &mWCov& mRec & mPrec &mCov &mWCov\\ 
    \hline
    ASIS~\cite{wang2019associatively}              & 35.1 & 46.6 & 34.7 & 41.6 & 35.3 & 48.8 & 35.1 & 42.3 & 0.2 & 2.2 & 0.4 & 0.7\\ 
    JSNet~\cite{zhao2020jsnet}              & 48.3 & 60.1 & 44.4 & 49.9 & \bf{48.5} &  \bf{64.4} & \bf{44.9} & \bf{51.0} & 0.2 & 4.3& 0.5 & 1.1 \\     
    PointGroup*~\cite{jiang2020pointgroup}  & 45.5 & 52.2 & 41.6 & 46.9 & 47.1 & 59.4 & 42.8 & 48.9 & 1.6 & 7.2 & 1.2 & 2.0 \\ 
    DyCo3D*~\cite{he2021dyco3d}             & 47.4 & 56.6 & 43.6 & 49.0 & 48.3 & 60.4 & 44.8 & 50.5 & 0.9 & 3.8& 1.2 & 1.5 \\ 
    HAIS*~\cite{chen2021hierarchical}       & 42.7 & 48.2 & 39.5 & 45.3 & 45.5 & 54.3 & 41.3& 47.8 & 2.8 & 6.1 & 1.8 & 2.5 \\ 
    \hline
\end{tabular}
\label{tab:results2}
\end{center}
\end{table*}



\section{Conclusion}
This paper presents a novel block merging algorithm that significantly improves the prediction accuracy compared to the reference merging method that is currently employed by all block-based instance segmentation techniques in the literature. The increased performance is obtained by enabling wrongly labelled points from past processed blocks to be corrected through label propagation. By doing so, instance overlap when processing blocks is not anymore required to achieve consistent results, which is the main limitation of the original method. Experiments consistently show that state-of-the-art performance is obtained when adopting the proposed block merging algorithm.   
For mean recall and precision, improvements up to 2.8\% and 7.2\% are obtained. For mean coverage and weighted coverage, the improvements add up to 1.8\% and 2.5\%, respectively. Visually, the additional performance is translated to less object confusion, ultimately leading to clearer and more refined instance segmentation. 

Conceptually, the proposed method will benefit all existing block-based instance segmentation methods in the literature. Investigating this aspect is left as topic of further investigation.


\section*{Acknowledgment}
This research is funded by Innoviris (BRGRD66-TORRES) and FWO (1S89420N).
\par DOI: www.doi.org/10.1109/DSP58604.2023.10167976

\bibliographystyle{IEEEtran}
\bibliography{refs}

\begin{thebibliography}{10}
\providecommand{\url}[1]{#1}
\csname url@samestyle\endcsname
\providecommand{\newblock}{\relax}
\providecommand{\bibinfo}[2]{#2}
\providecommand{\BIBentrySTDinterwordspacing}{\spaceskip=0pt\relax}
\providecommand{\BIBentryALTinterwordstretchfactor}{4}
\providecommand{\BIBentryALTinterwordspacing}{\spaceskip=\fontdimen2\font plus
\BIBentryALTinterwordstretchfactor\fontdimen3\font minus
  \fontdimen4\font\relax}
\providecommand{\BIBforeignlanguage}[2]{{%
\expandafter\ifx\csname l@#1\endcsname\relax
\typeout{** WARNING: IEEEtran.bst: No hyphenation pattern has been}%
\typeout{** loaded for the language `#1'. Using the pattern for}%
\typeout{** the default language instead.}%
\else
\language=\csname l@#1\endcsname
\fi
#2}}
\providecommand{\BIBdecl}{\relax}
\BIBdecl

\bibitem{jiang2020pointgroup}
L.~Jiang, H.~Zhao, S.~Shi, S.~Liu, C.-W. Fu, and J.~Jia, ``Pointgroup: Dual-set
  point grouping for 3d instance segmentation,'' in \emph{Proceedings of the
  IEEE/CVF Conference on Computer Vision and Pattern Recognition}, 2020, pp.
  4867--4876.

\bibitem{han2020occuseg}
L.~Han, T.~Zheng, L.~Xu, and L.~Fang, ``Occuseg: Occupancy-aware 3d instance
  segmentation,'' in \emph{Proceedings of the IEEE/CVF conference on computer
  vision and pattern recognition}, 2020, pp. 2940--2949.

\bibitem{he2021dyco3d}
T.~He, C.~Shen, and A.~van~den Hengel, ``Dyco3d: Robust instance segmentation
  of 3d point clouds through dynamic convolution,'' in \emph{Proceedings of the
  IEEE/CVF Conference on Computer Vision and Pattern Recognition}, 2021, pp.
  354--363.

\bibitem{chen2021hierarchical}
S.~Chen, J.~Fang, Q.~Zhang, W.~Liu, and X.~Wang, ``Hierarchical aggregation for
  3d instance segmentation,'' in \emph{Proceedings of the IEEE/CVF
  International Conference on Computer Vision}, 2021, pp. 15\,467--15\,476.

\bibitem{vu2022softgroup}
T.~Vu, K.~Kim, T.~M. Luu, T.~Nguyen, and C.~D. Yoo, ``Softgroup for 3d instance
  segmentation on point clouds,'' in \emph{Proceedings of the IEEE/CVF
  Conference on Computer Vision and Pattern Recognition}, 2022, pp. 2708--2717.

\bibitem{wang2018sgpn}
W.~Wang, R.~Yu, Q.~Huang, and U.~Neumann, ``Sgpn: Similarity group proposal
  network for 3d point cloud instance segmentation,'' in \emph{Proceedings of
  the IEEE conference on computer vision and pattern recognition}, 2018, pp.
  2569--2578.

\bibitem{wang2019associatively}
X.~Wang, S.~Liu, X.~Shen, C.~Shen, and J.~Jia, ``Associatively segmenting
  instances and semantics in point clouds,'' in \emph{Proceedings of the
  IEEE/CVF Conference on Computer Vision and Pattern Recognition}, 2019, pp.
  4096--4105.

\bibitem{zhao2020jsnet}
L.~Zhao and W.~Tao, ``Jsnet: Joint instance and semantic segmentation of 3d
  point clouds,'' in \emph{Proceedings of the AAAI Conference on Artificial
  Intelligence}, vol.~34, no.~07, 2020, pp. 12\,951--12\,958.

\bibitem{zhang2021point}
B.~Zhang and P.~Wonka, ``Point cloud instance segmentation using probabilistic
  embeddings,'' in \emph{Proceedings of the IEEE/CVF Conference on Computer
  Vision and Pattern Recognition}, 2021, pp. 8883--8892.

\bibitem{chen2022jspnet}
F.~Chen, F.~Wu, G.~Gao, Y.~Ji, J.~Xu, G.-P. Jiang, and X.-Y. Jing, ``Jspnet:
  Learning joint semantic \& instance segmentation of point clouds via feature
  self-similarity and cross-task probability,'' \emph{Pattern Recognition},
  vol. 122, p. 108250, 2022.

\bibitem{elich20193d}
C.~Elich, F.~Engelmann, T.~Kontogianni, and B.~Leibe, ``3d bird’s-eye-view
  instance segmentation,'' in \emph{German Conference on Pattern
  Recognition}.\hskip 1em plus 0.5em minus 0.4em\relax Springer, 2019, pp.
  48--61.

\bibitem{pham2019jsis3d}
Q.-H. Pham, T.~Nguyen, B.-S. Hua, G.~Roig, and S.-K. Yeung, ``Jsis3d: Joint
  semantic-instance segmentation of 3d point clouds with multi-task pointwise
  networks and multi-value conditional random fields,'' in \emph{Proceedings of
  the IEEE/CVF Conference on Computer Vision and Pattern Recognition}, 2019,
  pp. 8827--8836.

\bibitem{he2020instance}
T.~He, Y.~Liu, C.~Shen, X.~Wang, and C.~Sun, ``Instance-aware embedding for
  point cloud instance segmentation,'' in \emph{European Conference on Computer
  Vision}.\hskip 1em plus 0.5em minus 0.4em\relax Springer, 2020, pp. 255--270.

\bibitem{he2020learning}
T.~He, D.~Gong, Z.~Tian, and C.~Shen, ``Learning and memorizing representative
  prototypes for 3d point cloud semantic and instance segmentation,'' in
  \emph{European Conference on Computer Vision}.\hskip 1em plus 0.5em minus
  0.4em\relax Springer, 2020, pp. 564--580.

\end{thebibliography}

\end{document}